\pdfoutput=1

\documentclass[11pt]{article}

\usepackage[final]{acl}

\usepackage{times}
\usepackage{latexsym}

\usepackage[T1]{fontenc}

\usepackage[utf8]{inputenc}

\usepackage{microtype}

\usepackage{inconsolata}

\usepackage{graphicx}

\usepackage{multicol}

\usepackage{hyperref}

\title{Retcon - a Prompt-Based Technique for Precise Control of LLMs in Conversations}

\author{
  David Kogan \quad Sam Nguyen \\
  {\bf  Masanori Suzuki \quad Feiyang Chen} \AND
  Google
}

\begin{document}
\maketitle
\begin{abstract}

Recent advances in Large Language Models (LLMs) allow agents to execute complex natural language tasks. Many LLM applications, such as support agents, teaching assistants, and interactive bots, involve multi-turn conversations. However, it remains challenging to control LLMs in the context of such interactions, particularly when the LLM behavior needs to be adjustable over the course of the conversation. In this paper, we present Retcon, a few-shot prompting technique designed to provide turn-level control over LLMs in conversations. We then demonstrate that it performs significantly better than  zero-shot and traditional few-shot prompting.
\end{abstract}

\section{Introduction}

In the domain of conversational agents, a key ability is being able to adjust responses to meet desired conditions. For example, a support agent may be instructed to adjust its tone \cite{Balamurali2023SentimentAF}, a game character may be instructed to react to its simulated environment \cite{MatyasTheEO}, or a teaching agent may be instructed to adjust difficulty \cite{Ali2023SupportingSL}.

However, controlling agent responses with traditional techniques including zero-shot (prompting a model without examples) and few-shot (prompting a model with examples) can be difficult \cite{ZamfirescuPereira2023HerdingAC}, especially when the desired responses do not match the tone and content of prior turns in the conversation \cite{Gupta2024LLMTI} or when the conversation is more than a few turns long \cite{Yan2024RefuteBenchER}. While it is possible to improve on individual tasks using fine-tuning \cite{Xu2023WizardLMEL} or controllability frameworks \cite{Li2024FormalLLMIF}, such approaches are costly in both training effort and compute, and prompting is preferable in many real-world applications \cite{petrov2024when}.

There's therefore the need for a prompting technique that allows better controllability, but does not require fine-tuning an LLM.

In this work, we present Retcon, an iteration of few-shot for use in LLM conversations. We test it on a challenging conversational task and demonstrate that it performs better than zero-shot and traditional few-shot.

\section{Related Work}

The GPT paper \cite{brown2020languagemodelsfewshotlearners} demonstrated that few-shot prompting is an effective way to adapt LLMs to new tasks and achieve good performance. Since then, numerous works have been done to explore different prompting techniques to improve such results.

The most common prompting techniques are zero-shot prompting \cite{reynolds2021promptprogramminglargelanguage} and few-shot prompting, with few-shot performing better in many cases, particularly dependent on the number of few-shot examplars and their order \cite{lu2022fantasticallyorderedpromptsthem} \cite{liu2021makesgoodincontextexamples}. Considerable research has been done to identify and optimize the exact format of prompts for both these techniques \cite{yang2024largelanguagemodelsoptimizers} \cite{zhou2024selfdiscoverlargelanguagemodels} \cite{Bhandari2023ASO}.

Other techniques include reasoning \cite{wei2023chainofthoughtpromptingelicitsreasoning} and planning \cite{zhou2023leasttomostpromptingenablescomplex} \cite{Li2023APS}, though these typically require substantially more compute time or larger models. Fine-tuning is another established way to improve results \cite{Xu2023WizardLMEL} \cite{Shin2023PromptEO}, but is much more expensive than prompting and is not accessible to most LLM users \cite{trad2024prompt} \cite{Xu2023WizardLMEL}.

Notably, most of the approaches mentioned above target the improvement of a single response LLM, e.g. question answering. Retcon focuses specifically on per-turn controllability within a multi-turn conversation.

\section{Preliminary Knowledge}

The most common prompting techniques for LLM tasks are zero-shot and few-shot prompting \cite{schulhoff2024promptreportsystematicsurvey}. With zero-shot, the LLM is only given instructions, and with few-shot, the LLM is additionally given concrete examples.

Consider a conversational task where the LLM is asked to respond to conversation C at turn k, with some goal G (e.g. G = "cheerfulness: 0.5"). The goal is specific to that turn, so we annotate it as G\textsubscript{k}. To differentiate this conversation from examples given to the model, we'll annotate it with f for final, so conversation C\textsubscript{f}, at turn k\textsubscript{f} with goal G\textsubscript{f,k\textsubscript{f}}.

There are some prior number of turns in the conversation C\textsubscript{f}, comprised of turns T\textsubscript{f} (T\textsubscript{f,1}...T\textsubscript{f,k\textsubscript{f}-1}) and as this is a live conversation, these turns are not known before execution time. The number of prior turns may be zero for the case where the LLM is expected to start the conversation.

There are additionally x pregenerated static conversations C\textsubscript{n} each composed of some number of turns T\textsubscript{n} (T\textsubscript{n,1}..T\textsubscript{n,k\textsubscript{n}}) where for the last turn T\textsubscript{n,k\textsubscript{n}} of each conversation, G\textsubscript{n,k\textsubscript{n}} was precomputed, such that we know that T\textsubscript{n,k\textsubscript{n}} is a good response for the goal G\textsubscript{n,k\textsubscript{n}}.

There is also optionally a static instruction overview O that can be provided at the start of each conversation.

With traditional few-shot prompting, the prompt is constructed as follows:

\begin{quote}
O\\
\color{blue}
T\textsubscript{1,1}\\
T\textsubscript{1,2}\\
\color{black}
...\\
\color{blue}
T\textsubscript{1,k\textsubscript{1}-1}\\
\color{red}
I(G\textsubscript{1,k\textsubscript{1}})\\
\color{violet}
T\textsubscript{1,k\textsubscript{1}}\\
\color{black}
O\\
\color{blue}
T\textsubscript{2,1}\\
T\textsubscript{2,2}\\
\color{black}
...\\
\color{blue}
T\textsubscript{x,k\textsubscript{x}-1}\\
\color{red}
I(G\textsubscript{x,k\textsubscript{x}})\\
\color{violet}
T\textsubscript{x,k\textsubscript{x}}\\
\color{black}
O\\
\color{blue}
T\textsubscript{f,1}\\
T\textsubscript{f,2}\\
\color{black}
...\\
\color{blue}
T\textsubscript{f,k\textsubscript{f}-1}\\
\color{red}
I(G\textsubscript{f,k\textsubscript{f}})
\color{black}
\end{quote}

There are other permutations of prompt ordering \cite{Mao2023PromptPR}, for example, I(G\textsubscript{n,k\textsubscript{n}}) can be placed at the start of the conversation instead of the end, but this structure is taken as representative.

The key observation is that for each precomputed conversation C\textsubscript{n}, the LLM is given exactly one example of how to respond, at turn T\textsubscript{n,k\textsubscript{n}}. Increasing the number of examples to improve quality \cite{liu2021makesgoodincontextexamples} requires authoring new example conversations, which can be difficult and expensive \cite{zhao2021calibrateuseimprovingfewshot} and significantly increases the context length, and therefore computation cost and latency \cite{vaswani2023attentionneed}.

Zero-shot is simply a special case of few-shot where the number of example conversations is zero:

\begin{quote}
O\\
\color{blue}
T\textsubscript{f,1}\\
T\textsubscript{f,2}\\
\color{black}
...\\
\color{blue}
T\textsubscript{f,k\textsubscript{f}-1}\\
\color{red}
I(G\textsubscript{f,k\textsubscript{f}})
\color{black}
\end{quote}

For full prompt examples, see \ref{sec:prompt_examples}.

\begin{figure*}[t]
  \includegraphics[width=0.48\linewidth]{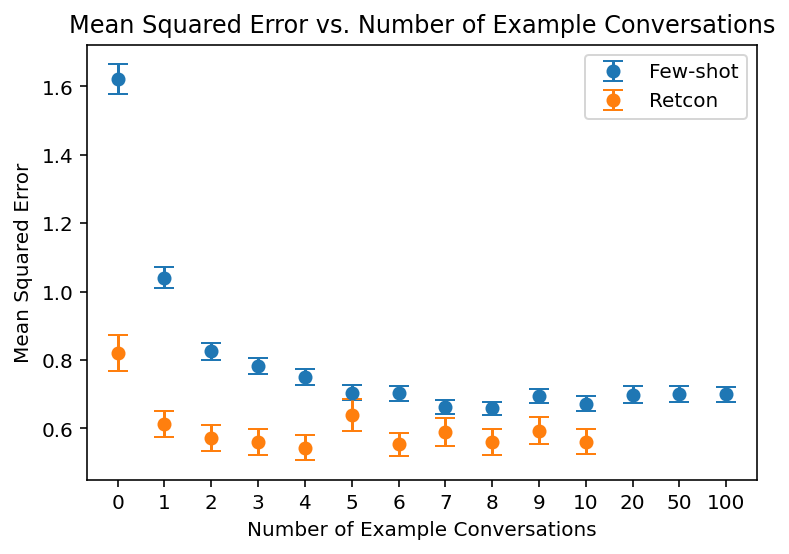} \hfill
  \includegraphics[width=0.48\linewidth]{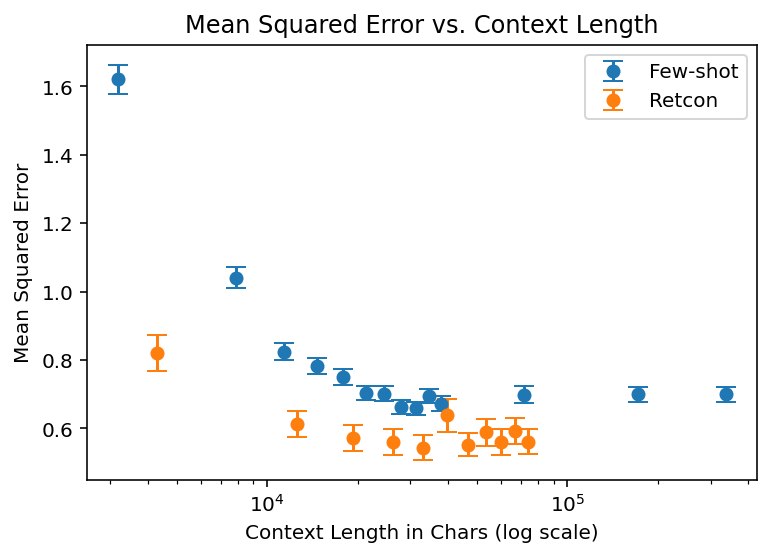}
  \caption {Mean squared error for regular few-shot and Retcon few-shot, vs number of examples on the left, and vs total context length in characters on the right, with 95\% confidence intervals. The far left point on both graphs corresponds to zero-shot (0 examples). The y-axis is MSE on the CEFR scale where each interval (e.g. A1-A2) is one unit.}
  \label{fig:results}%
\end{figure*}

\section{Retcon}

\subsection{Overview}

Retcon is a few-shot technique that makes each turn in a conversation serve as an example to the LLM, rather than each conversation. The examples include the turns of the current, ongoing conversation.

A Retcon prompt is authored by rewriting the conversation history to inject an instruction before each conversation turn. This rewrite is applied both within example conversations, and to the current ongoing conversation (hence "Retcon", after "retconning" in serialized fiction, since the conversation with the user is being rewritten on the fly).

The technique creates an additional system requirement, which is to have an evaluation function E(T) that evaluates the desired goal for a given text. (e.g. E(T\textsubscript{n,k\textsubscript{m}}) = "The measured cheerfulness of turn T\textsubscript{n,k\textsubscript{m}}"). Such evaluation functions are typical for evaluation and training, but in this case must be integrated into the serving path. 

\subsection{Creating a Retcon Prompt}

The prompt is constructed similarly to a regular few-shot prompt, but instead of instructions injected at the end of each example conversation, they are injected before every other turn (to simulate instructions given only to the LLM) or before every turn (to simulate instructions given to everyone) as follows:

\begin{quote}
O\\
\color{red}
I(E(T\textsubscript{1,1}))\\
\color{violet}
T\textsubscript{1,1}\\
\color{red}
I(E(T\textsubscript{1,2}))\\
\color{violet}
T\textsubscript{1,2}\\
\color{black}
...\\
\color{red}
I(E(T\textsubscript{1,k\textsubscript{1}}))\\
\color{violet}
T\textsubscript{1,k\textsubscript{1}}\\
\color{black}
O\\
\color{red}
I(E(T\textsubscript{2,1}))\\
\color{violet}
T\textsubscript{2,1}\\
\color{black}
...\\
\color{red}
I(E(T\textsubscript{x,k\textsubscript{x}}))\\
\color{violet}
T\textsubscript{x,k\textsubscript{x}}\\
\color{black}
O\\
\color{red}
I(E(T\textsubscript{f,1}))\\
\color{violet}
T\textsubscript{f,1}\\
\color{black}
...\\
\color{red}
I(E(T\textsubscript{f,k\textsubscript{f}-1}))\\
\color{violet}
T\textsubscript{f,k\textsubscript{f}-1}\\
\color{red}
I(G\textsubscript{f,k\textsubscript{f}})
\color{black}
\end{quote}

Every turn is preceded by an instruction, creating an example for the LLM. The number of examples given to the LLM is ($\sum_{n=1} ^{x} k\textsubscript{n}) + k\textsubscript{f} - 1$ (number of turns, including the current conversation), compared to x for few-shot (number of conversations).

This does substantially increase the length of the context compared to few-shot for the same number of example conversations, but accuracy increases even accounting for this, as shown in section \ref{sec:experiment}.

For complete examples of what these prompts look like, see appendix \ref{sec:prompt_examples}.

\section{Experiment}

\label{sec:experiment}

\subsection{Existing Benchmarks}

While there are many benchmarks for evaluating LLMs in conversations \cite{hendrycks2021measuringmassivemultitasklanguage, zheng2023judgingllmasajudgemtbenchchatbot}, they are predominantly evaluating an agent's ability to answer questions or understand users. We were unable to identify an example of a dataset designed to evaluate how well a model adapts to instructions within a conversation. We therefore created our own, as detailed below.

\subsection{Experiment Setup}

For our experiment, we tested zero-shot, regular few-shot, and Retcon few-shot against the task of responding to a conversation using a specific language difficulty level, as could be used to help a user learning English. For the difficulty scale, we used the Common European Framework of Reference (CEFR) scale.

We used identical prompt texts, with the control variables being the techniques used and the number of example conversations provided.

The overall prompt O instructed the model to pretend to be an English instructor and have a conversation with a learner, adjusting the complexity of responses as directed, as well as giving a refresher of the CEFR scale. (\ref{sec:prompt_overall}). Instructions I(G) were given as directives to respond with one of the CEFR levels (\ref{sec:prompt_instruction}). Example responses were formatted in JSON including the CEFR difficulty (\ref{sec:prompt_instruction}) and the structured schema Gemini API was used to ensure the model produced output in the same format. Turns were labeled as either "ASSISTANT" or "STUDENT", and additional labels indicated to the model when a new conversation was beginning and who would go first. See Appendix (\ref{sec:prompt_examples}) for full example prompts for each of zero-shot, few-shot and Retcon.

For the data set, we manually authored 20 conversations of 20 turns each, on a variety of topics (\ref{sec:conversation}). A manual effort was made to author turns representing a variety of difficulty levels from A1 (beginner) to C2 (advanced). Half of the conversations (10) were randomly chosen to be used as examples, and the other half (10) were used for eval. The same split was used in every case. 

To maintain consistency when comparing different models, we used the same static conversations as the prior turns for each model. In future work, it would be worth comparing the models in live conversations as well, though direct evaluation will be challenging due to conversations diverging over time.

For eval, we called Gemini via API, using the model Gemini Pro 1.1. For each test condition, we ran 2520 queries asking for a conversation response: 2x for each combination of eval conversation (10 conversations), number of prior turns (21, including 0 prior turns), and requested difficulty level on the CEFR scale (6: A1, A2, B1, B2, C1, C2).

For the evaluation function, we used a Bert-based difficulty measuring model trained using the techniques developed by \cite{devlin2019bertpretrainingdeepbidirectional,arase2022cefrbasedsentencedifficultyannotation,kogan2025acecefrdatasetautomated}. An English learning language expert manually validated the model and established that it has an MSE of < 0.4 on the scale of A1-C2 where each interval (e.g. A1 to A2) is measured as 1 unit. This evaluation function was used for instructions as well as for measuring response error, to ensure alignment between examples given to the LLM, and the evaluation of its response.

Few-shot and Retcon were evaluated with 0 to 10 example conversations. Note that at 0 example conversations, few-shot is just zero-shot. For 1 to 10 examples, conversations were chosen randomly without replacement from the example pool of 10. For each few-shot example, a random conversation length k\textsubscript{n} was chosen between 0 and 20, to provide examples of varying conversation lengths. (If k\textsubscript{n} is constant, few-shot only performs well if k\textsubscript{n} = k\textsubscript{f}.)

We gave instructions to Retcon every other turn, so with 10 example conversations used, Retcon has 100+ example turns, compared to only 10 for few-shot. Therefore, for further comparison, we tested few-shot with 20, 50, and 100 examples, reusing the same 10 example conversations, but ending at different turns.

Example model outputs can be see in \hyperref[sec:responseexamples]{Appendix C}.

\subsection{Results}

Retcon significantly outperformed few-shot at every example conversation count other than one outlier where the confidence intervals overlapped (Figure \ref{fig:results}). The best Retcon result was MSE of 0.544±.036, compared to few-shot 0.659±.020.

It's notable that Retcon prompts are substantially longer than few-shot prompts for the same number of example conversations, due to more instruction text injected. Since LLM cost is proportional to the size of the context, we additionally measured average context length versus mean squared error for each example count (Figure \ref{fig:results}). With this comparison as well, Retcon is significantly better aside from the same outlier.

Few-shot did not outperform Retcon even when given a comparable number of turn examples or more. With 100 example conversations (100 annotated example turns), few-shot MSE was 0.7±0.044, compared to Retcon with 8 example conversation (80-100 annotated example turns, depending on current conversation length) MSE of 0.56±0.038. Both techniques achieved their best results before the maximum number of examples: Retcon's best results were with 4 example conversations, and few-shot's best results were with 8.

It's also notable also that with 0 examples, zero-shot has almost double the error of Retcon, with MSE 1.621±0.043 compared to 0.821±0.052. This is because every turn of the current conversation provides Retcon with an example, even if no prior example conversations are available.

\section{Conclusion}

Retcon performs better than few-shot and zero-shot for adjusting text difficulty, for a large range of example counts and prompt lengths. Retcon also reaches better overall performance, with fewer examples, than the best performance of few-shot.

\section{Future Work}

Future work is desirable to understand more precisely the mechanism by which Retcon operates. Retcon has three distinct effects: an increase in the number of example turns, an increase in the density of examples, and a closer proximity of examples to the final instruction. It is likely that all three contribute to improved performance, and verifying this and measuring the impact of each will be useful for determining when and how to apply the technique. Clarifying the underlying mechanisms may also reduce or eliminate the need for an integrated serving-time eval function.

Further research into what kinds of tasks Retcon works well on, and how it compares to other techniques is also desirable. We have only tested Retcon against other prompting techniques, and evaluating it relative to fine-tuning and chain-of-thought would also be of some interest.

For the results above, we also only compared the models using the same exact prior turns for each one. It would be useful to also compare the models in live conversation, though effort will need to be made to ensure comparability as the conversations diverge.

\section{Limitations}

We evaluated the technique only on English, using one model, on one task. The effects may not translate to other languages, other tasks, or other LLMs. Measurement across a variety of conditions is needed to establish whether Retcon performs consistently better than few-shot, if there are cases where it performs worse, or if there are cases where it performs comparably while incurring significant additional complexity. This is particularly unclear because Retcon performs better than few-shot even with comparable numbers of instruction examples, indicating that there are multiple factors contributing to its success.

A key limitation of the technique itself is the need to integrate an evaluation model into the serving flow. This may be simple for some tasks (e.g. detecting whether a word is present) and challenging for others (e.g. measuring emotion). This may be prohibitive for developers who lack the ability to access or create such a models.

Also the creation of example and eval conversations can be a challenging obstacle in many cases. While two of the authors of this paper had the background to create our data sets, and we were able to directly author them, these are not always readily available skills. In many cases, in order to create examples, vendor labor is used, which can raise ethical concerns about fair compensation for such work, and appropriate subsequent usage of the results.

Finally, because Retcon provides an improved fine-grained control over LLMs in conversation, it increases the risk of abuse by malicious actors using LLMs. For example, a company could use Retcon to inject subtle advertisements into its support agent, without making the end user aware of it. As with any technique designed to prompt or control AI-driven systems, efforts should be made to align the user needs with the design of the system, and to provide transparency about the system's behavior. It would be productive to create legal frameworks about the behavior and transparency of such systems, so as to reduce the chances of such malicious applications.

\bibliography{custom}

\appendix

\section{Prompts}

\subsection{Introductory}
\label{sec:prompt_overall}

\begin{quote}
You are an expert instructor of English as a second language. Help your student practice English conversational skills. Respond, adjusting the difficulty of your responses on the CEFR scale, as instructed.\\
As a reminder, the CEFR scale is the Common European Framework of Reference. It's used to evaluate the ability of second language learners. Here are the levels:\\
A1: Student is a complete beginner. Use only the most basic simple words and extremely short sentences with simple construction.\\
A2: Student has been learning for a year, but is still a beginner. Use simple words and short sentences.\\
B1: Student has been learning for two years, and is an early intermediate. Use common words and simple sentences.\\
B2: Student has been learning for three years, and can understand normal conversation. Use normal words and typical sentences.\\
C1: Student has been learning for four years, and is becoming advanced. Use complex vocabulary and sentence structure.\\
C2: Student has been learning for more than five years and is an expert in the language. Use extremely complex vocabulary and sentence structure.\\
Follow instructions in parentheses, but do not respond to the instructions.
\end{quote}

\subsection{Instruction}

\label{sec:prompt_instruction}

Instruction prompt, where <target> is one of "A1", "A2", "B1", "B2", "C1", "C2".

\subsubsection{Conversation with Zero Prior Turns}

\begin{quote}
Your task: Begin a conversation as ASSISTANT. Your conversation turn must have an English language difficulty of exactly <target> on the CEFR scale.
\end{quote}

\subsubsection{Conversation with One or More Prior Turns}

\begin{quote}
Your task: Respond as ASSISTANT. Your conversation turn must have an English language difficulty of exactly <target> on the CEFR scale.
\end{quote}

\subsection{Response Format}

\label{sec:prompt_response}

Response format where <target> is one of "A1", "A2", "B1", "B2", "C1", "C2" and <text> is the response text.

\begin{verbatim}
{
  "text_difficulty": "<target>",
  "text": "<text>"
}    
\end{verbatim}

\subsection{Example Prompts}

\label{sec:prompt_examples}

Color coding is added for readability, and is not provided to the LLM.

\subsubsection{Example Zero-shot Prompt}

<Introductory from \ref{sec:prompt_overall}>\\
(START OF CONVERSATION)\\
(STUDENT will go first.)\\
\color{blue}
STUDENT: Did you bring matches for the campfire?\\
ASSISTANT: I'm not sure. Were they on my list?\\
STUDENT: I think I forgot to put them on either list. They were so obvious.\\
\color{red}
(Your task: Respond as ASSISTANT. Your conversation turn must have an English language difficulty of exactly B1 on the CEFR scale.)
\color{black}

\subsubsection{Example Few-shot Prompt}

<Introductory from \ref{sec:prompt_overall}>\\
Follow the following examples\\
EXAMPLE 0:\\
<Introductory from \ref{sec:prompt_overall}>\\
(START OF CONVERSATION)\\
(ASSISTANT will go first.)\\
\color{blue}
ASSISTANT: Did you hear about the time an art piece was lost in transit and the gallery displayed just the pedestal?\\
STUDENT: What do you mean?\\
ASSISTANT: I fail to recall where exactly, but this was many years ago. An artist posted a sculpture to a museum, but the package split up in mid-transit and only the pedestal delivered.\\
STUDENT: And they displayed it?\\
\color{red}
(Your task: Respond as ASSISTANT. Your conversation turn must have an English language difficulty of exactly C2 on the CEFR scale.)\\
\color{violet}
\{\\
  "text\_difficulty": "C2",\\
  "text": "Yes, and while that's understandable at first, the artist quickly realized their misunderstanding, and let them know, but by that point it was too late, and the quote unquote sculpture was seen by a ton of visitors."\\
\}\\
\color{black}
EXAMPLE 1:\\
<Introductory from \ref{sec:prompt_overall}>\\
(START OF CONVERSATION)\\
(STUDENT will go first.)\\
\color{blue}
STUDENT: It's hot!\\
\color{red}
(Your task: Respond as ASSISTANT. Your conversation turn must have an English language difficulty of exactly C1 on the CEFR scale.)\\
\color{violet}
\{\\
  "text\_difficulty": "C1",\\
  "text": "I concur, it beggars belief. I'm sweating through all my clothes, and it's barely the end of spring."\\
\}\\
\color{black}
YOUR TASK:\\
<Introductory from \ref{sec:prompt_overall}>\\
(START OF CONVERSATION)\\
(STUDENT will go first.)\\
\color{blue}
STUDENT: Did you bring matches for the campfire?\\
ASSISTANT: I'm not sure. Were they on my list?\\
STUDENT: I think I forgot to put them on either list. They were so obvious.\\
\color{red}
(Your task: Respond as ASSISTANT. Your conversation turn must have an English language difficulty of exactly A1 on the CEFR scale.)
\color{black}

\subsubsection{Example Retcon Prompt}

<Introductory from \ref{sec:prompt_overall}>\\
Follow the following examples\\
EXAMPLE 0:\\
<Introductory from \ref{sec:prompt_overall}>\\
(START OF CONVERSATION)\\
(ASSISTANT will go first.)\\
\color{red}
(Your task: Begin a conversation as ASSISTANT. Your conversation turn must have an English language difficulty of exactly B1 on the CEFR scale.)\\
\{\\
  "text\_difficulty": "B1",\\
  "text": "Which do you like better, your phone or your computer?"\\
\}\\
\color{blue}
STUDENT: Well, I'm upon my phone twenty-four seven, and I'm obligated to use my computer to acquire money, so I'd hazard both are pretty terrible for me as a human being. What sort of choice do you expect?\\
\color{red}
(Your task: Respond as ASSISTANT. Your conversation turn must have an English language difficulty of exactly B2 on the CEFR scale.)\\
\color{violet}
\{\\
  "text\_difficulty": "B2",\\
  "text": "Easy, which one would you rather live without?"\\
\}\\
\color{blue}
STUDENT: Do I have a job?\\
\color{red}
(Your task: Respond as ASSISTANT. Your conversation turn must have an English language difficulty of exactly B1 on the CEFR scale.)\\
\color{violet}
\{\\
  "text\_difficulty": "B1",\\
  "text": "Can you afford not to?"\\
\}\\
\color{blue}
STUDENT: No.\\
\color{red}
(Your task: Respond as ASSISTANT. Your conversation turn must have an English language difficulty of exactly B1 on the CEFR scale.)\\
\color{violet}
\{\\
  "text\_difficulty": "B1",\\
  "text": "Then yes, you still have to work."\\
\}\\
\color{blue}
STUDENT: Would I be permitted to just go out and buy another one, either immediately, or in a week or a year, or, would I be coerced into spending my entire life without acquiring the one I forgo?\\
\color{red}
(Your task: Respond as ASSISTANT. Your conversation turn must have an English language difficulty of exactly C2 on the CEFR scale.)\\
\color{violet}
\{\\
  "text\_difficulty": "C2",\\
  "text": "Designate one and endure perpetually."\\
\}\\
\color{blue}
STUDENT: Well, then obviously computer. It's hard to do anything if I can't work.\\
\color{red}
(Your task: Respond as ASSISTANT. Your conversation turn must have an English language difficulty of exactly C2 on the CEFR scale.)\\
\color{violet}
\{\\
  "text\_difficulty": "C2",\\
  "text": "I see your point. Perhaps my inquiry was somewhat lacking in rationality."\\
\}\\
\color{blue}
STUDENT: Exactly. Which would you rather have, your head or your body?\\
\color{red}
(Your task: Respond as ASSISTANT. Your conversation turn must have an English language difficulty of exactly B1 on the CEFR scale.)\\
\color{violet}
\{\\
  "text\_difficulty": "B1",\\
  "text": "Is the cell phone the head, or the computer?"\\
\}\\
\color{blue}
STUDENT: I think the computer is the body, since it does all the work. And the cell phone is the head, because it just mindlessly scrolls all day.\\
\color{red}
(Your task: Respond as ASSISTANT. Your conversation turn must have an English language difficulty of exactly B1 on the CEFR scale.)\\
\color{violet}
\{\\
  "text\_difficulty": "B1",\\
  "text": "You're funny. How about for a week?"\\
\}\\
\color{blue}
STUDENT: Let's talk about something else. How's your kid doing in school?\\
\color{red}
(Your task: Respond as ASSISTANT. Your conversation turn must have an English language difficulty of exactly B2 on the CEFR scale.)\\
\color{violet}
\{\\
  "text\_difficulty": "B2",\\
  "text": "Oh, she's great! She's just finishing up her senior year. She got accepted in all the schools she applied to!"\\
\}\\
\color{blue}
STUDENT: Great! Where is she going?\\
\color{red}
(Your task: Respond as ASSISTANT. Your conversation turn must have an English language difficulty of exactly B2 on the CEFR scale.)\\
\color{violet}
\{\\
  "text\_difficulty": "B2",\\
  "text": "Yeah, she's going to Berkely, can you believe it?"\\
\}\\
\color{blue}
STUDENT: Indeed, the sense of accomplishment must be palpable. Seeing your efforts come to fruition is truly gratifying. Congratulations are in order!\\
\color{black}
YOUR TASK:\\
<Introductory from \ref{sec:prompt_overall}>\\
(START OF CONVERSATION)\\
(STUDENT will go first.)\\
\color{blue}
STUDENT: Did you bring matches for the campfire?\\
\color{red}
(Your task: Respond as ASSISTANT. Your conversation turn must have an English language difficulty of exactly B1 on the CEFR scale.)\\
\color{violet}
\{\\
  "text\_difficulty": "B1",\\
  "text": "I'm not sure. Were they on my list?"\\
\}\\
\color{blue}
STUDENT: I think I forgot to put them on either list. They were so obvious.\\
\color{red}
(Your task: Respond as ASSISTANT. Your conversation turn must have an English language difficulty of exactly A1 on the CEFR scale.)
\color{black}

\section{Example Conversation}

\label{sec:conversation}
- Which do you like better, your phone or your computer?\\
- Well, I'm upon my phone twenty-four seven, and I'm obligated to use my computer to acquire money, so I'd hazard both are pretty terrible for me as a human being. What sort of choice do you expect?\\
- Easy, which one would you rather live without?\\
- Do I have a job?\\
- Can you afford not to?\\
- No.\\
- Then yes, you still have to work.\\
- Would I be permitted to just go out and buy another one, either immediately, or in a week or a year, or, would I be coerced into spending my entire life without acquiring the one I forgo?\\
- Designate one and endure perpetually.\\
- Well, then obviously computer. It's hard to do anything if I can't work.\\
- I see your point. Perhaps my inquiry was somewhat lacking in rationality.\\
- Exactly. Which would you rather have, your head or your body?\\
- Is the cell phone the head, or the computer?\\
- I think the computer is the body, since it does all the work. And the cell phone is the head, because it just mindlessly scrolls all day.\\
- You're funny. How about for a week?\\
- Let's talk about something else. How's your kid doing in school?\\
- Oh, she's great! She's just finishing up her senior year. She got accepted in all the schools she applied to!\\
- Great! Where is she going?\\
- Yeah, she's going to Berkeley, can you believe it?\\
- Indeed, the sense of accomplishment must be palpable. Seeing your efforts come to fruition is truly gratifying. Congratulations are in order!

\section{Example Responses}
\label{sec:responseexamples}

Example responses comparing Few-shot(2) to Retcon(2) for the same conversation. Difficulty is measured on the CEFR scale, with A1=1 and C2 = 6.

\subsection{Prior Conversation}

- Did you bring matches for the campfire? \\
- I'm not sure. Were they on my list? \\
- I think I forgot to put them on either list. They were so obvious. \\
- Oh no. Did you bring anything else that might help? \\
- I don't think so, but perhaps we can make do in some other fashion. Have you ever learned how to begin a campfire without using matches or anything fancy like that? \\
- I think you can rub two sticks together, right? \\
- Yes, but I read somewhere that's very difficult. What about a magnifying glass, do you have one? \\
- Why? \\
- I don't know, to look at insects? \\
- Sorry. No. \\
- So what do we do? \\
- We could hike back to the car, and go stop by a gas station to get some. What do you think? \\
- That's a two hour hike each way, it will be well past dark by that point.

\subsection{Responses}

Example responses for few-shot are shown in \hyperref[fig:fewshotresponses]{Figure 2} and for retcon in  \hyperref[fig:retconresponses]{Figure 3}. The prompt of each model includes two example conversations. Each model was run once for each combination of inputs, and the first response generated is shown.

\begin{figure*}
\begin{tabular}{ p{1.5cm} p{11cm} p{1.5cm} c }
 Target Difficulty &
 Response &
 Actual Difficulty &
 Error \\
 \hline

 1 & 
 No walk. & 
 1.0 &
 0 \\ 

 2 & 
 Maybe we can ask another group if they have some matches. & 
 3.4 &
 1.4 \\ 

 3 & 
 We could try to find some dry tinder and use a fire starter. Or, we could try to use the sun's rays with a piece of glass. & 
 4.8 &
 1.8 \\ 

 4 & 
 True, that's not ideal. We could also try to gather some dry tinder and leaves, and see if we can start a fire with a flint. Do you have a pocket knife or anything sharp? & 
 4.7 &
 0.7 \\ 

 5 & 
 In this predicament, we find ourselves at an impasse. Hiking back to the car in the waning light would be ill-advised. However, we could potentially utilize natural resources in our immediate vicinity to ignite the campfire. For instance, if we could find some dry tinder, we might be able to start a fire using friction. & 
 5.9 &
 0.9 \\ 

 6 & 
 Well, in that case, we might have to get creative. There are several ways to start a fire without matches or a magnifying glass, although some of them require a bit of skill and practice. One method is the fire plow, which involves rapidly sawing a groove into a piece of soft wood using a harder piece of wood. Another is the hand drill, where a pointed stick is twirled between the palms while bearing down on a piece of tinder. Both techniques generate heat through friction, eventually creating an ember that can be nurtured into a flame. Of course, there's always the option of foraging for natural fire starters like dry leaves, twigs, and pine needles to build a tinder bundle. With a little patience and ingenuity, we should be able to get a fire going without having to make the long trek back to the car. & 
 5.6 &
 -0.4 \\ 
\end{tabular}
  \caption {Example responses from few-shot(2) model for different levels of target difficulty.}
  \label{fig:fewshotresponses}%
  \end{figure*}

\begin{figure*}
\begin{tabular}{ p{1.5cm} p{11cm} p{1.5cm} c }
 Target Difficulty &
 Response &
 Actual Difficulty &
 Error \\
 \hline

 1 & 
 Dark? & 
 1.0 &
 0 \\ 

 2 & 
 Not good. & 
 1.14 &
 -0.9 \\ 

 3 & 
 That's true. We should probably figure out another option. & 
 4.2 &
 1.2 \\ 

 4 & 
True, but we need to be realistic about the options we have. I'm up for a hike if you are, or we can take our chances trying to start a fire without matches. & 
 4.7 &
 0.7 \\ 

 5 & 
 I understand. Let us explore alternative solutions. The sun is setting soon, so time is of the essence. & 
 4.7 &
 -0.3 \\ 

 6 & 
 Under these suboptimal circumstances, our only feasible option rests in invoking the power of ingenuity. We must seek alternative means to initiate combustion without the aid of conventional tools or implements. Consider the phenomena of friction, the production of heat through kinetic energy. We could construct a fire plow, or perhaps employ flint knapping techniques to generate sparks. The key lies in our ability to manipulate the materials at our disposal, maximizing their potential and unlocking the secrets of ancient fire-starting methods. & 
 6.0 &
 0.0 \\ 
\end{tabular}
  \caption {Example responses from retcon(2) model for different levels of target difficulty.}
  \label{fig:retconresponses}%
  \end{figure*}
  
\end{document}